\documentclass[11pt,letterpaper]{article}

\usepackage[letterpaper]{geometry}

\usepackage{tacl}

\usepackage{times}

\usepackage{latexsym}

\usepackage{amsmath}

\usepackage{multirow}

\usepackage{url}

\usepackage{amssymb}

\usepackage{colortbl}

\definecolor{very-likely}{RGB}{92,172,238}
\definecolor{likely}{RGB}{198,226,255}
\definecolor{plausible}{RGB}{238,233,233}
\definecolor{tech-plausible}{RGB}{255,193,193}
\definecolor{impossible}{RGB}{240,128,128}

\definecolor{word}{RGB}{179,226,205}
\definecolor{sentence}{RGB}{203,213,232}
\makeatletter

\newcommand{\@BIBLABEL}{\@emptybiblabel}

\newcommand{\@emptybiblabel}[1]{}
\newcommand{\ph}{\phantom{*}}

\makeatother

\usepackage[hidelinks]{hyperref}

\usepackage{amsmath} 

\usepackage{hyperref}
\hypersetup{colorlinks=true,citecolor=blue}

\usepackage[utf8]{inputenc}

\usepackage{cleveref}
\crefname{section}{§}{§§}
\Crefname{section}{§}{§§}
\Crefname{figure}{Fig}{}
\Crefname{algorithm}{Alg.}{}

\usepackage[usenames,dvipsnames,svgnames,table]{xcolor}
\usepackage{soul}
\newcommand{\Note}[1]{}
\renewcommand{\Note}[1]{\hl{[#1]}}

\definecolor{lgreen}{RGB}{179,226,205}

\newcommand{\Reviewer}[1]{}
\newcommand{\Slot}{\_\_\_\_}

\usepackage[draft]{todonotes}   

\usepackage{mdframed} 
\usepackage{enumitem} 
\usepackage{graphicx} 
\usepackage{wasysym} 
\usepackage{booktabs}
\usepackage{verbatimbox}
\usepackage{caption}
\usepackage{subcaption}

\title{Ordinal Common-sense Inference}

\author{Sheng Zhang\\
  Johns Hopkins University\\
  {\tt zsheng2@jhu.edu} \\\And
  Rachel Rudinger\\
  Johns Hopkins University\\
  {\tt rudinger@jhu.edu} \\\AND
  Kevin Duh\\
  Johns Hopkins University\\
  {\tt kevinduh@cs.jhu.edu} \\\And
  Benjamin Van Durme\\
  Johns Hopkins University\\
  {\tt vandurme@cs.jhu.edu} \\}

\date{}

\begin{document}

\maketitle

\begin{abstract}



  Humans have the capacity to draw common-sense inferences from
  natural language: various things that are likely but not certain to
  hold based on established discourse, and are rarely stated
  explicitly.  We propose an evaluation of automated common-sense
  inference based on an extension of recognizing textual entailment:
  predicting ordinal human responses on the subjective likelihood of an
  inference holding in a given context.  We describe a framework for
  extracting common-sense knowledge from corpora, which is then used to
  construct a dataset for this ordinal entailment task.
  We train a neural sequence-to-sequence model on this dataset,
  which we use to score and generate possible inferences.
  Further, we annotate subsets of previously established
  datasets via our ordinal annotation protocol in order to then
  analyze the distinctions between these and what we have constructed.

\end{abstract}

\section{Introduction}


\begin{quote}
  {\small\emph{We use words to talk about the world. Therefore, to understand what words mean, we must have a prior explication of how we view the world.} --~\newcite{hobbs87}}
\end{quote}
Researchers in Artificial Intelligence and (Computational) Linguistics
have long-cited the requirement of common-sense knowledge in language understanding.%
\footnote{\newcite{schank75}: \emph{It has been apparent ... within ... natural language understanding ... that the eventual limit to our solution ... would be our ability to characterize world knowledge.}} %
This knowledge is viewed as a key component in filling in the gaps between the telegraphic style of natural language statements: we are able to convey considerable information in a relatively sparse channel, presumably owing to a partially shared model at the start of any discourse.%
\footnote{\newcite{mccarthy59}: \emph{a program has common sense if it automatically deduces for itself a sufficiently wide class of immediate consequences of anything it is told and what it already knows.}}
    \Reviewer{B: I believe that the reference to Clark 1975
  and bridging is quite wrong, since bridging refers to a
  much more specific inference process, namely the inference
  of non-identity semantic relations between nominal
  expression in adjacent discourse.}

\begin{figure}[t!]
  \begin{center}
  \small
  \begin{tabular}{|p{0.9\columnwidth}|}
      \hline
  {\cellcolor{very-likely}Sam bought a new clock $\leadsto$ {\bf The clock runs}}\\ 
  {\cellcolor{likely}Dave found an axe in his garage $\leadsto$ {\bf A car is parked in the garage}}\\
  {\cellcolor{plausible}Tom was accidentally shot by his teammate in the army $\leadsto$ {\bf The teammate dies}}\\
  {\cellcolor{tech-plausible}Two friends were in a heated game of checkers $\leadsto$ {\bf A person shoots the checkers}}\\
  {\cellcolor{impossible}My friends and I decided to go swimming in the ocean $\leadsto$ {\bf The ocean is carbonated}}\\
  \hline
  \end{tabular}
  \caption{\small Examples of common-sense inference ranging from {\em very likely}, {\em likely}, {\em plausible}, {\em technically possible}, to {\em impossible}. \label{fig:possibilistic_examples}}
\end{center}
\end{figure}

Common-sense inference -- inferences based on common-sense knowledge
-- is \emph{possibilistic}: things everyone more or less would expect
to hold in a given context, but without the necessary strength of logical
entailment.\footnote{E.g., many of the \emph{bridging} inferences of
  \newcite{clark75} make use of common-sense knowledge, such as the
  following example of ``Probable part'': \emph{I walked into the
    room. The windows looked out to the bay.} To resolve the
  definite reference \emph{the windows}, one needs to
  know that \emph{rooms have windows} is probable.} %
Because natural language corpora exhibit human reporting bias~\cite{gordon13}, systems that derive knowledge exclusively from such
corpora may be more accurately considered models of language, rather than of the world~\cite{rudinger15}.
Facts such as ``A person walking into a room is very likely to be
\textit{blinking} and \textit{breathing}''
are usually unstated in text, so their real-world
likelihoods do not align to language model probabilities.\footnote{For
  further background see discussions by \newcite{vandurme-thesis},
  \newcite{gordon13}, \newcite{rudinger15} and
  \newcite{visual-reporting-bias}.}  
We would like to have systems capable of, e.g., reading a sentence that
describes a real-world situation and inferring how likely other 
statements about that situation are to hold true \textit{in the real world}.
This capability is subtly but crucially distinct from the ability to
predict other sentences reported in the same text, as a language model may be trained to do.

We therefore propose a model of knowledge acquisition based on first
deriving possibilistic statements from text: as the relative frequency
of these statements suffers the mentioned reporting bias, we then
follow up with human annotation of derived examples. 
Since we initially are uncertain about the real-world likelihood of the
derived common-sense knowledge holding in any particular context,
we pair it with various grounded context and present to humans for  their
own assessment.
As these examples vary in assessed
plausibility, we propose the task of \emph{ordinal} common-sense
inference, which embraces a wider set of natural conclusions arising
from language comprehension (see \Cref{fig:possibilistic_examples}).




In what follows, we describe prior efforts in
common-sense and textual inference (\S\ref{sec:oti}).
We then state our position on how ordinal common-sense inference should be defined (\S\ref{sec:position}),
and detail our own framework for large-scale
extraction and abstraction, along with a crowdsourcing protocol for
assessment (\S\ref{sec:framework}).
This includes a novel neural model for \emph{forward generation} of textual inference statements.
Together these methods are applied to contexts derived from various
prior textual inference resources, resulting in the JHU Ordinal Common-sense
Inference (JOCI) corpus, a large collection of diverse common-sense inference
examples, judged to hold with varying levels of
subjective likelihood (\S\ref{sec:corpus}).  We provide baseline results (\S\ref{sec:prediction})
for prediction on the JOCI corpus.\footnote{The JOCI corpus is released freely at:
\url{http://decomp.net/}.}
\Reviewer{C: What is missing is a compelling argument for what they would gain if other people reimplement similar methods over their own data.}




\section{Background}\label{sec:oti}
\Reviewer{B: I do not feel that the relationship between common sense
  knowledge, common sense inference, and ordinal inferences
  is clarified sufficiently in the paper. As I mentioned
  above, the paper starts out talking about common sense
  knowledge, but then moves on to talk about common sense
  inference without really connecting the two topics. From
  that perspective, I don't even see the relevance of the
  discussion of common sense mining in Section 2 -- none
  of the rest of the paper appears to relate to that.}
\paragraph{Mining Common Sense} Building large collections of common-sense knowledge can be done
manually via professionals~\cite{hobbsUnpublished93}, but at
considerable cost in terms of time and
expense~\cite{miller1995wordnet,lenat1995cyc,baker-fillmore-lowe:1998:ACLCOLING,friedland2004project}.
Efforts have pursued volunteers \cite{singhAAAISpSymp02,havasiRANLP07}
and games with a purpose \cite{chklovskiKCAP03}, but are still left
fully reliant on human labor.  Many have pursued automating the
process, such as in expanding lexical
hierarchies~\cite{hearstCOLING92,snow06}, constructing inference
patterns~\cite{lin01,berant2011}, reading reference
materials~\cite{richardsonACL98,suchanekWWW07}, mining search engine
query logs~\cite{pascaIJCAI07}, and most relevant here: abstracting
from instance-level predications discovered in descriptive
texts~\cite{schubert2002can,liakataCOLING02,clark2003knowledge,bankoKCAP07}.
In this article we are concerned with knowledge mining for purposes of
seeding a text generation process (constructing common-sense inference
examples).  \Reviewer{C: I strongly doubt that the authors have a
  basis for claiming that all previous efforts assume that the
  resultant collection of explicit knowledge is a useful artifact. I
  agree that previous workers may have assumed something tending
  towards this, but the sentence needs reformulating to make a more
  modest and appropriate claim.}

\paragraph{Common-sense Tasks} Many textual inference
tasks have been designed to require some degree of common-sense 
knowledge, e.g., the Winograd Schema Challenge discussed
by~\newcite{levesque2011}. The data for these tasks are either
smaller, carefully constructed evaluation sets by professionals,
following efforts like the \textsc{FraCaS} test
suite~\cite{cooper1996using}, or they rely on crowdsourced
elicitation~\cite{snli:emnlp2015}. Crowdsourcing is scalable, but
elicitation protocols can lead to biased responses unlikely to contain
a wide range of possible common-sense inferences: humans can generally
agree on the plausibility of a wide range of possible inference pairs,
but they are not likely to generate them from an initial
prompt.\footnote{\newcite{mcrae2005}: \emph{For example, features such
    as $<$is larger than a tulip$>$ or $<$moves faster than an
    infant$>$, although logically possible, do not occur in [human
      responses] [...] Although people are capable of verifying that a
    $<$dog is larger than a pencil$>$}.}

The construction of SICK (Sentences Involving Compositional Knowledge)
made use of existing paraphrastic sentence pairs (descriptions by
different people of the same image), which were modified through a
series of rule-based transformations then judged by
humans~\cite{sick}.  As with SICK, we rely on humans only for judging
provided examples, rather than elicitation of text.  Unlike SICK, our
generation is based on a process targeted specifically
at common sense (see \S\ref{sec:gwke}).

\Reviewer{C: the claim that the author's generation
uses a "more extensive" process than SICK is worryingly vague, and could be
reformulated to be less so.}

\paragraph{Plausibility}
\Reviewer{B: the exclusive motivation of ordinal inference
  as a generalization of textual inference misses an equally
  important related concept, namely the notion of plausibility,
  which has been a touchstone of psycholinguistics for quite
  a while (see e.g. the work by McRae who is cited in the
  article) and on which there is also a considerable amount
  of computational modeling work. Ordinal inference may be
  seen as a move from an unconditional notion of plausibility
  ("absolute plausibility") to a conditional notion of
  plausibility ("plausibility in context"). This angle, which
  I believe offers an interesting perspective on the task,
  as well as a completely different handle on modeling, is
  currently completely missing from the paper.}

Researchers in psycholinguistics have explored a notion of
plausibility in human sentence processing, where, for instance,
arguments to predicates are intuitively more or less ``plausible'' as
fillers to different thematic roles, as reflected in human reading
times.  For example, \newcite{mcrae98} looked at manipulations such as:
\vspace{-1mm}
\begin{center}
  \begin{tabular}{p{0.9\columnwidth}}
    (a) The boss hired by the corporation was perfect for the job. \\
    (b) The applicant hired by the corporation was perfect for the job.
  \end{tabular}
\end{center}

\noindent where the plausibility of a \emph{boss} being the agent --
as compared to patient -- of the predicate \emph{hired} might be
measured by looking at delays in reading time in the words following
the predicate; this measurement is then contrasted with the timing 
observed in the same positions in (b).\footnote{This notion of thematic plausibility
  is then related to the notion of verb-argument selectional
  preference~\cite{zernikCOLING92,resnikARPA93,clarkEMNLP99}, and
  sortal (in)correctness~\cite{thomasonJoPL72}.}

Rather than measuring according to predictions such as human
reading times, here we ask annotators explicitly to judge plausibility
on a 5-point ordinal scale (See \S\ref{sec:position}).  Further, our effort might be described in
this setting as \emph{conditional} plausibility,\footnote{Thanks to
  anonymous reviewer for this connection.} where plausibility
judgments for a given sentence are expected to be dependent on
preceding context.  Further exploration of conditional plausibility is
an interesting avenue of potential future work, perhaps through the
measurement of human reading times when using prompts derived from our
ordinal common-sense inference examples.
Computational modeling of (unconditional) semantic
plausibility has been explored by those such as
\newcite{pado09probabilistic}, \newcite{erk2010} and
\newcite{sayeed2015exploration}.

\paragraph{Textual Entailment}
A multi-year source of textual inference examples were generated under
the Recognizing Textual Entailment (RTE) Challenges, introduced by \newcite{dagan2006}:
\begin{small}
\begin{quote}
\emph{We say that T entails H if, typically, a human reading T would
  infer that H is most likely true. This somewhat informal definition
  is based on (and assumes) common human understanding of language as
  well as common background knowledge.}
\end{quote}
\end{small}

\Reviewer{C: The first sentence of the section labeled textual entailment has a number
agreement problem, and, in any case, I don't know what a multi-year source
is supposed to be.}
This definition strayed from the more strict notion of entailment as
used by linguistic semanticists, such as those involved with
\textsc{FraCaS}. While~\newcite{giampiccolo2008} extended binary RTE
with an ``unknown" category, the entailment community has primarily focussed on issues such as paraphrase and
monotonicity, such as captured by the Natural Logic implementation of \newcite{macartney2007}.

Language understanding in context is not only understanding the entailments
of a sentence, but also the plausible inferences of the sentence, i.e. the
new posterior on the world after reading the sentence. A new
sentence in a discourse is almost never entailed by another sentence in the
discourse, because such a sentence would add no new information. In order
to successfully process a discourse, there needs to be some understanding of
what new information can be possibly or plausibly added to the discourse. 
Collecting sentence pairs with ordinal entailment connections is potentially
useful for improving and testing these language understanding capabilities
that would be needed by algorithms for applications like storytelling.

\newcite{garrette2011integrating} and
\newcite{beltagy2016representing}
treated textual entailment as
probabilistic logical inference in Markov Logic Networks~\cite{richardson2006markov}.
But the notion of probability in their entailment task
has a subtle distinction from our problem of common-sense inference.
The probability of being an
entailment given by a probabilistic model trained for a binary classification
(being an entailment or not), is not necessarily the same as the likelihood
of an inference being true. For example:
\begin{center}
  \begin{tabular}{p{0.9\columnwidth}}
    T:\,A person flips a coin.\\
    H:\,That flip comes up heads.
  \end{tabular}
\end{center}
No human reading $T$ should infer that $H$ is true.  A model trained
to make ordinal predictions should say: ``plausible, with probability
1.0'', whereas a model trained to make binary entailed/not-entailed
predictions should say: ``not entailed, with probability 1.0''.
The following example exhibits the same property:
\begin{center}
  \begin{tabular}{p{0.9\columnwidth}}
    T: An animal eats food.\\
    H: A person eats food.
  \end{tabular}
\end{center}
Again, with high confidence, $H$ is plausible; and, with high confidence, it is also not entailed.

\paragraph{Non-entailing Inference}
Of the various non-``entailment'' textual inference tasks, a few are
most salient here.  \newcite{agirre2012} piloted a \emph{Textual
  Similarity} evaluation which has been refined in subsequent
years: systems produce scalar values corresponding to predictions of
how similar the meaning is between two provided sentences.  E.g.,~the following pair from SICK was judged very similar (4.2
out of 5), while also being a contradiction: \emph{There is no biker
  jumping in the air} and \emph{A lone biker is jumping in the air}.
The ordinal approach we advocate for relies on a graded notion,
like textual similarity.
\Reviewer{C: I don't understand what the last sentence of the first paragraph of the
section on Non-entailing inference is trying to claim.}

The Choice of Plausible Alternative (COPA)
task~\cite{roemmele_choice_2011} was a reaction to RTE, similarly
motivated to probe a system's ability to understand inferences that
are not strictly entailed: a single context was
provided, with two alternative inferences, and a system had to judge
which was more plausible.  The COPA dataset was manually elicited, and
is not large: we discuss this data further in \S\ref{sec:corpus}.

The Narrative Cloze task~\cite{chambersACL08} requires a system to
score candidate inferences as to how likely they are to appear in a
document that also included the provided context. 
Many such inferences are then not strictly \emph{entailed} by the context.
Further, the cloze task gives the benefit of being able to generate
very large numbers of examples automatically by simply occluding parts
of existing documents and asking a system to predict what is missing.
The LAMBADA dataset~\cite{paperno-EtAl:2016:P16-1} is akin to our strategy
for automatic generation followed by human filtering, but for cloze examples.
As our concern is with inferences that are often true but never stated
in a document, this approach is not viable here.
The ROCStories corpus~\cite{mostafazadeh-EtAl:2016:N16-1} elicited a more
``plausible'' collection of documents in order to retain the narrative cloze
in the context of common-sense inference. The ROCStories corpus can be viewed
as an extension of the idea behind the COPA corpus, done at a larger scale
with crowdsourcing, and with multi-sentence contexts; we consider this
dataset in \S\ref{sec:corpus}.
\Reviewer{C:  when Mostafazadeh et al are stated as having elicited a more "everyday"
collection of documents, it is unclear to the reader what the real
difference from what Chambers and Jurafsky did is. Presumably their
documents are less "everyday". What does this mean, and what are the
consequences?}

Alongside the narrative cloze, \newcite{pichotta2016} made use of a
5-point Likert scale (very likely to very unlikely) as a secondary
evaluation of various script induction techniques.  While they were
concerned with measuring their ability to generate very likely
inferences, here we are interested in generating a wide swath of
inference candidates, including those that are impossible.


\section{Ordinal Common-sense Inference}
\Reviewer{B: I am wondering about the motivation for ordinal
  inferences, in the sense of their usefulness. It seems to
  me that compared to textual inference, where there at least
  a strong prior for "soundness" of inferences in the logical
  sense, ordinal inference is arguably a much weaker concept,
  and so it is in more need to show its empirical usefulness.
  I appreciate that adding some kind of extrinsic evaluation
  is a tall order, but I currently just do not feel comfortable
  with the level of motivation given for ordinal inferences.
  One very interesting angle, I feel, is exactly the relationship
  to plausibility, which may potentially provide such a
  motivation, plus a link to datasets that can be used for
  comparison and thus provide a better understanding of the
  novel concept.}
  \Reviewer{C: The goal stated at the outset of section 3 is not motivated, and is
certainly not attached to any definite task.}
\label{sec:position}

Our goal is a system that can perform speculative, common-sense
inference as part of understanding language.  Based on the observed
shortfalls of prior work, we propose the notion of
\emph{Ordinal Common-sense Inference (OCI)}. OCI embraces the notion of
\newcite{dagan2006}, in that we are concerned with human judgements of
epistemic modality.\footnote{Epistemic modality: the likelihood that (some
    aspect of) a certain state of affairs is/has been/will be true (or false)
    in the context of the possible world under consideration.}

\begin{quote}
    {\small\emph{%
As agreed by many linguists, modality in natural language is a continuous
category, but speakers are able to map areas of this axis into discrete values
\cite{lyons1977semantics,horn1989natural,de1997interaction}
} --~\newcite{sauri2009factbank}}
\end{quote}



\begin{figure}[h!]
\centering
    \includegraphics[width=0.45\textwidth]{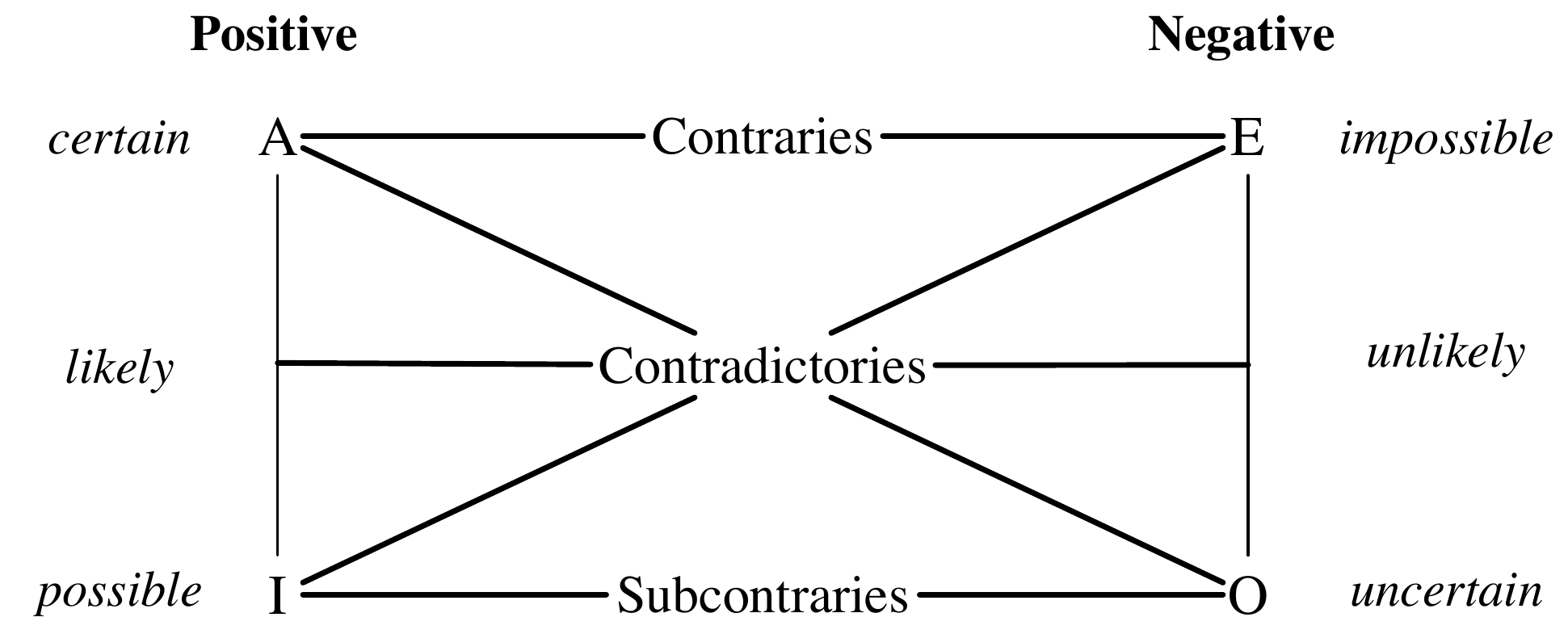}
    \caption[]{SO for epistemic modals
    \cite{sauri2009factbank}.\footnotemark\label{fig:SO}}
\end{figure}

According to \newcite{horn1989natural}, there are two scales of epistemic modality which
differ in polarity (positive vs.\ negative polarity):
$\langle$\emph{certain, likely, possible}$\rangle$
and $\langle$\emph{impossible, unlikely, uncertain}$\rangle$.
The Square of Opposition (SO) (\Cref{fig:SO}) illustrates the logical
relations holding between values in the two scales.
Based on their logical relations, we can make a set of exhaustive epistemic modals:
$\langle$\emph{very likely, likely, possible, impossible}$\rangle$,
where $\langle$\emph{very likely, likely, possible}$\rangle$
lie on a single, positive Horn scale, and \emph{impossible}, a complementary
concept from the corresponding negative Horn scale, completes the set.
In this paper, we further replace the value \emph{possible} by the more fine-grained values
(\emph{technically possible} and \emph{plausible}).
This results in a 5-point scale of likelihood:
$\langle$\emph{very likely, likely, plausible, technically possible, impossible}$\rangle$.
The OCI task definition directly embraces subjective likelihood on such an ordinal scale.
Humans are presented with a context $\mathcal{C}$ and asked
whether a provided hypothesis $\mathcal{H}$ is \textit{very likely},
\textit{likely}, \textit{plausible},
\textit{technically possible}, or \textit{impossible}.
Furthermore, an important part of this process is the generation of $\mathcal{H}$ by automatic methods,
which seeks to avoid the elicitation bias of many prior works.
\Reviewer{C: Did you manage to avoid elicitation bias, as claimed in the last sentence
of the same section?  Are there metrics which can be used to collect
evidence for success or failure in this endeavor.}
\vspace{-2mm}
\footnotetext{``Contradictories":
            exhaustive and mutually exclusive conditions. ``Contraries":
            non-exhaustive and mutually exclusive. ``Subcontraries":
        exhaustive and non-mutually exclusive.}

\section{Framework for collecting OCI corpus}
\label{sec:framework}
We now describe our framework for collecting ordinal common-sense
inference examples. It is natural to collect this data in two stages.
In the first stage (\S\ref{sec:agci}), we automatically generate
inference candidates given some context. 
We propose two broad approaches using either general world knowledge
or neural methods.
In the second stage (\S\ref{sec:ordinal_label_annotation}), we annotate these candidates with ordinal labels.

\subsection{Generation of Common-sense Inference Candidates}

\label{sec:agci}

\subsubsection{Generation based on World Knowledge}
\label{sec:gwke}
\Reviewer{B: why did you design the algorithm in
  the way you did?}
Our motivation for this approach was first introduced by
\newcite{schubert2002can}:
\begin{small}
    \begin{quote}
        \emph{There is a largely untapped source of general knowledge in
            texts, lying at a level beneath the explicit assertional content.
            This knowledge consists of relationships implied to be possible
            in the world, or, under certain conditions, implied to be normal
        or commonplace in the world.}
    \end{quote}
\end{small}

\begin{figure}[h!]
\centering
    \includegraphics[width=0.45\textwidth]{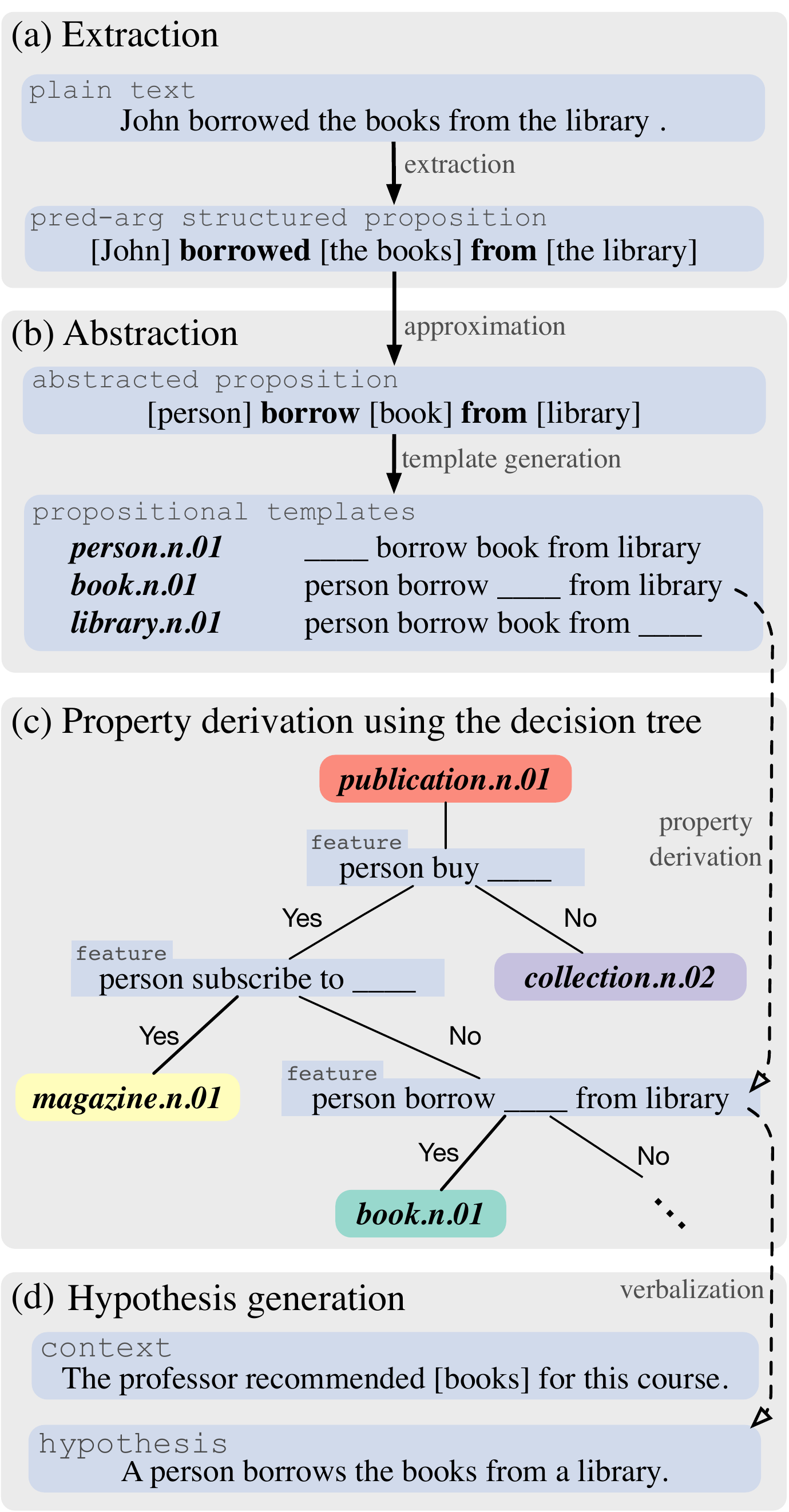}
\caption{Generating common-sense inferences based on general world knowledge.
\label{fig:knowledge_derivation}}
\end{figure}

Following \newcite{schubert2002can} and \newcite{van2008open}, we define an approach
for abstracting over explicit assertions derived from corpora, leading
to a large-scale collection of general possibilistic statements. As
shown in \Cref{fig:knowledge_derivation}, this approach generates
common-sense inference candidates in four steps:
(a) extracting propositions with predicate-argument structures from texts,
(b) abstracting over propositions to generate templates for concepts,
(c) deriving properties of concepts via different strategies,
and (d) generating possibilistic hypotheses from contexts.


\noindent\textbf{(a) Extracting propositions}: First we extract a large set of propositions with predicate-argument
structures from noun phrases and clauses, under which general world
presumptions often lie.
To achieve this goal, we use PredPatt\footnote{\footnotesize \url{https://github.com/hltcoe/PredPatt}}~\cite{white2016universal},
which defines a framework of interpretable, language-neutral
predicate-argument extraction patterns from Universal Dependencies~\cite{deMarneffe2014universal}.
\Cref{fig:knowledge_derivation} (a) shows an example extraction.

We use the Gigaword corpus~\cite{parker2011english} for extracting propositions
as it is a comprehensive text archive.
There exists a version containing automatically generated
syntactic annotation \cite{ferraro-2014-concretely}, which bootstraps large-scale knowledge
extraction.  We use
PyStanfordDependencies\footnote{\footnotesize \url{https://pypi.python.org/pypi/PyStanfordDependencies}}
to convert constituency parses to depedency parses, from which we extract
structured propositions.

%
%
%

\noindent\textbf{(b) Abstracting propositions}:
\label{subsec:abstraction}
In this step, we abstract the propositions into a more general form.
This involves lemmatization, stripping inessential modifiers and conjuncts,
and replacing specific arguments with generic types.\footnote{For example (using English
glosses of the logical representations), abstraction of ``\emph{a long, dark
corridor}" would yield ``\emph{corridor}"; ``\emph{a small office at the end of a
long dark corridor}" would yield ``\emph{office}"; and ``\emph{Mrs. MacReady}"
would yield ``\emph{person}". See~\newcite{schubert2002can} for detail.}
This method of abstraction
often yields general presumptions about the world.
To reduce noise from predicate-argument extraction,
we only keep 1-place and 2-place predicates after abstraction.
\Reviewer{C: Why does the restriction to 1 and 2 place predicates reduce noise and
avoid the data-sparsity problem? How many 3 place and higher predicates are
there anyway?}



We further generalize individual arguments to concepts by attaching
semantic-class labels to them.
Here we choose WordNet~\cite{miller1995wordnet} noun synsets\footnote{In order to avoid too
    general senses, we set cut points at the depth of 4
\cite{pantel2007isp} to truncate the hierarchy and consider all 81,861 senses
below these points.}
as the semantic-class set.
When selecting the correct sense for an argument, we adopt a fast and
relatively accurate method:
always taking the first sense which is usually the most commonly used sense
\cite{suchanekWWW07,pasca2008turning}.
By doing so, we attach 84 million abstracted propositions with senses,
covering 43.7\% (35,811/81,861) of WordNet noun senses.



Each of these WordNet senses, then, is associated with a set of abstracted propositions.
The abstracted propositions are turned into templates by replacing the sense's corresponding argument with a placeholder, similar to
\newcite{vandurme-michalak-schubert:2009:EACL} (see \Cref{fig:knowledge_derivation} (b)).
We remove any template associated with a sense if it occurs less than two times for that sense, leaving 38 million unique templates.






\noindent\textbf{(c) Deriving properties via WordNet}:
At this step, we want to associate with each WordNet sense a set of possible
properties. We employ three strategies.



The first strategy is to use a decision tree to pick out highly discriminative properties for each WordNet sense.
Specifically, for each set of co-hyponyms,\footnote{Senses sharing a hypernym
with each other are called co-hyponyms (e.g., \textit{book.n.01},
\textit{magazine.n.01} and \textit{collections.n.02} are co-hyponyms of
\textit{publication.n.01}).} we train a decision tree using the associated
templates as features.
For example, in \Cref{fig:knowledge_derivation} (c), we train a decision tree over the co-hyponyms of \textit{publication.n.01}.
Then the template ``person subscribe to~\Slot'' would be selected as a property of \textit{magazine.n.01}, and the template ``person borrow~\Slot~from library'' for \textit{book.n.01}.
The second strategy selects the most frequent templates associated with each sense as properties of that sense.
The third strategy uses WordNet ISA relations to derive new properties of senses.
E.g.~for the sense \textit{book.n.01} and its hypernym
\textit{publication.n.01}, we generate a property ``\Slot~be publication''.

\noindent\textbf{(d) Generating hypotheses}:
As shown in \Cref{fig:knowledge_derivation} (d), given a discourse context
\cite{tanenhaus80}, we first extract an
argument of the context, then
select the derived properties for the argument.
Since we don't assume any specific sense for the argument,
these properties could come from any of its candidate senses.
We generate hypotheses by replacing the placeholder in the
selected properties with the argument, and verbalizing the properties.\footnote{\@
We use the pattern.en module
({\footnotesize \url{http://www.clips.ua.ac.be/pages/pattern-en}}) for verbalization,
which includes determining plurality of the argument, adding proper articles,
and conjugating verbs.}
\Reviewer{C: At the end of 4.1.1, what is etc. ? Too vague.}

\subsubsection{Generation via Neural Methods}
\Reviewer{B: the description of the second method (the neural S2S
  model) is so short as to be almost contentless. What is the
  architecture of the model? How did you set the hyperparameters?
  What are the final values of the hyperparameters? The current
  description does not enable anyone to reproduce these results.}

\label{subsec:neural}


\Reviewer{C:  SNLI and ROCStories are referred to in 4.1.2, but introduces only in 5.1.}
In addition to the knowledge-based methods described above, we also adapt a neural sequence-to-sequence model \cite{vinyals15,bahdanau14translate} to generate inference candidates given contexts.
The model is trained on sentence pairs labeled ``entailment'' from the SNLI corpus~\cite{snli:emnlp2015} (train).
Here, the SNLI ``premise'' is the input (context $\mathcal{C}$), and the SNLI ``hypothesis'' is the output (hypothesis $\mathcal H$).

We employ two different strategies for \textit{forward generation} of inference candidates given any context.
The sentence-prompt strategy uses the entire sentence in the context as an input, and generates output using greedy decoding.
The word-prompt strategy differs by using only a single word from the context as input.
This word is chosen in the same fashion as the step (d) in generation based on world knowledge, i.e. an argument of the context.
The second approach is motivated by our hypothesis that providing only a
single word context will force the model to generate a hypothesis that
generalizes over the many contexts in which that word was seen, resulting
in more common-sense-like hypotheses, as in \Cref{Fig:seq2seq_example}.
We later present the full context and decoded hypotheses to crowdsource workers for annotation.

\begin{figure}[h!]
  \begin{center}
  \small
  \begin{tabular}{|p{0.9\columnwidth}|}
      \hline
      {dustpan $\leadsto$ \textbf{a person is cleaning.}}\\ 
      \hline
      {a boy in blue and white shorts is sweeping with a broom and dustpan. $\leadsto$ \textbf{a young man is holding a broom.}}\\
      \hline
  \end{tabular}
  \caption{\small Examples of sequence-to-sequence hypothesis generation from single-word and full-sentence inputs.\label{Fig:seq2seq_example}}
\end{center}
\end{figure}






\subsubsection*{Neural Sequence-to-Sequence Model}

Neural sequence-to-sequence models learn to map variable-length input
sequences to variable-length output sequences, as a conditional probability
of output given input.
For our purposes, we want to learn the conditional probability of an
hypothesis sentence, $\mathcal{H}$, given a context sentence, $\mathcal{C}$,
i.e., $P(\mathcal{H}|\mathcal{C})$.


The sequence-to-sequence architecture consists of two components: an encoder and a decoder.
The encoder is a recurrent neural network (RNN) iterating over input tokens
(i.e., words in $\mathcal{C}$), and the decoder is another RNN iterating over
output tokens (words in $\mathcal{H}$).
The final state of the encoder, $\textbf{h}_\mathcal{C}$, is passed to the decoder as its initial state.
We use a three-layer stacked LSTM (state size 512) for both the encoder and decoder RNN cells, with independent parameters for each.
We use the LSTM formulation of \newcite{hochreiter1997long} as summarized in \newcite{vinyals15}.

The network computes $P(\mathcal{H}|\mathcal{C})$:
\begin{equation}
P(\mathcal{H}|\mathcal{C}) = \prod\limits_{t=1}^{\text{len}(\mathcal{H})}
p(w_t|w_{<t},\mathcal{C})
\end{equation}
where $w_t$ are the words in $\mathcal{H}$.
At each time step, $t$, the successive conditional probability is computed from the LSTM's current hidden state:

\begin{equation}
p(w_t|w_{<t},\mathcal{C}) \propto \text{exp}(\textbf{v}_{w_t}\cdot \textbf{h}_t)
\label{eqn:cond}
\end{equation}
where $\textbf{v}_{w_t}$ is the embedding of word $w_t$ from its corresponding row in the output vocabulary matrix, $V$ (a learnable parameter of the network), and $\textbf{h}_t$ is the hidden state of the decoder RNN at time $t$.
In our implementation, we set the vocabulary to be all words that appear in the training data at least twice, resulting in a vocabulary of size 24,322.


This model also makes use of an attention mechanism.\footnote{See \newcite{vinyals15} for full details.}
An attention vector, $attn_t$, is concatenated with the LSTM hidden state at time $t$ to form the hidden state, $\textbf{h}_t$, from which output probabilities are computed (Eqn. \ref{eqn:cond}).
This attention vector is a weighted average of the hidden states of the encoder, $h_{1\leq i\leq \text{len}(\mathcal{C})}$:

\begin{equation}
\begin{aligned}
u_i^t &= v^T\text{tanh}(W_1h_i + W_2h_t)\\
a_i^t &= \text{softmax}(u_i^t)\\
attn_t &= \sum\limits_{i=1}^{\text{len}(\mathcal{C})}a_i^t h_i
\label{eqn:attn}
\end{aligned}
\end{equation}
where vector $v$ and matrices $W_1$, $W_2$ are parameters.

The network is trained via backpropagation on the cross-entropy loss of the observed sequences in training.
A sampled softmax is used to compute the loss during training, while a full
softmax is used after training to score unseen $(\mathcal{C},\mathcal{H})$
pairs, or generate an $\mathcal{H}$ given a $\mathcal{C}$.
Generation is performed via beam search with a beam size of 1; the highest probability word is decoded at each time step and fed as input to the decoder at the next time step until an end-of-sequence token is decoded.

\subsection{Ordinal Label Annotation} \label{sec:ordinal_label_annotation}
\Reviewer{B: Pertaining to both generation methods, I would like some discussion of the results of the generation in terms of sampling: can you say
  anything about the resulting corpus of inferences? Since the
  sentences (at least the consequents, if that term can still be applied)
  are not naturally occurring (at least not in the KR-based method),
  it is unclear what kind of a sample the resulting corpus represents.
  Since this corpus is supposed to be a training corpus, and almost all
  statistical methods make an i.i.d. assumption, this is a crucial point
  about the validity of the corpus.
  You do marginally touch on that point on Section 4.2, but I do not
  believe that this discussion is sufficient; by filtering out yet more
  examples it becomes even less clear what the properties of the
  resulting dataset are.}
In this stage, we turn to human efforts to annotate common-sense inference candidates with ordinal
labels.  The annotator is given a context, and then is asked to assess
the likelihood of the hypotheses being true.
These context-hypothesis pairs are annotated with one of the five labels:
\textit{very likely}, \textit{likely}, \textit{plausible},
\textit{technically possible}, and \textit{impossible}, corresponding
to the ordinal values of $\{$5,4,3,2,1$\}$ respectively.

In the case that the hypotheses in the
inference candidates do not make sense, or have grammatical errors, judges
can provide an additional label, \textit{NA}, so that we can filter
these candidates in post-processing. The combination of generation of common-sense inference candidates with human filtering seeks to avoid
the problem of elicitation bias.

\section{JOCI Corpus}
\label{sec:corpus}
We now describe in depth how we created the JHU Ordinal Common-sense Inference
(JOCI) corpus.  The main part of the corpus consists
of contexts chosen from SNLI~\cite{snli:emnlp2015} and ROCStories~\cite{mostafazadeh-EtAl:2016:N16-1}, paired with
hypotheses generated via methods described in
\S\ref{sec:agci}. These pairs are then annotated with ordinal labels
using crowdsourcing (\S\ref{sec:ordinal_label_annotation}).  We also
include context-hypothesis pairs directly taken from SNLI and other
corpora (e.g., as premise-hypothesis pairs), and re-annotate them with
ordinal labels.

\subsection{Data sources for Context-Hypothesis Pairs}

In order to compare with existing inference corpora, we choose
contexts from two resources: (1) the first sentence in the sentence pairs of
the SNLI corpus which are captions from the Flickr30k
corpus~\cite{young2014image}, and (2) the first sentence in the stories of the
ROCStories corpus.

We then collect candidates of automatically generated common-sense inferences
(AGCI) against these contexts.
Specifically, in the SNLI train set, there are over 150K
different first sentences, involving 7,414 different arguments according to
predicate-argument extraction. We randomly choose 4,600 arguments.
For each argument, we sample one first sentence that has the argument, and collect candidates of AGCI against this as context.
We also do the same generation for the SNLI development set and test set.
We also collect candidates of AGCI against randomly sampled first sentences in
the ROCStories corpus. Collectively, these pairs and their ordinal labels
(to be described in \cref{sec:crowdsourced_ordinal_annotation}) make up the
main part of the JOCI corpus.
The statistics of this subset are shown in \Cref{tab:joci_stat} (first five rows).

For comprehensiveness, we also produced ordinal labels on $(\mathcal{C},\mathcal{H})$ pairs directly drawn from existing corpora.
For SNLI, we randomly select 1000 contexts (premises) from the SNLI train set.
Then, the corresponding hypothesis is one of the entailment, neutral, or contradiction hypotheses taken from SNLI.
For ROCStories, we defined $\mathcal{C}$ as the first sentence of the story, and $\mathcal{H}$ as the second or third sentence.
For COPA, $(\mathcal{C},\mathcal{H})$ corresponds to premise-effect.
The statistics are shown in the bottom rows of Table~\ref{tab:joci_stat}.
\Reviewer{C:  I had difficulty in understanding what was meant by "the
AGCI subset"? Is this a subset you made?}


\begin{table}[h!]
\centering
\resizebox{0.48\textwidth}{!}{%
\begin{tabular}{|c|c|c|c|}
\hline
Subset Name & \# pairs & Context Source & Hypothesis Source \\ \hline\hline
\multirow{5}{*}{\parbox[c][1em][c]{0.14\textwidth}{\centering AGCI \\
\footnotesize against SNLI/ROCStories}} & 22,086 & SNLI-train & AGCI-\textsc{wk} \\ \cline{2-4}
& 2,456 & SNLI-dev & AGCI-\textsc{wk} \\ \cline{2-4}
 & 2,362 & SNLI-test & AGCI-\textsc{wk} \\ \cline{2-4}
 & 5,002 & ROCStories & AGCI-\textsc{wk} \\ \cline{2-4}
 & 1,211 & SNLI-train & AGCI-\textsc{nn} \\ \hline\hline
\multirow{3}{*}{SNLI} & 993 & SNLI-train & SNLI-entailment \\ \cline{2-4}
 & 988 & SNLI-train & SNLI-neutral \\ \cline{2-4}
 & 995 & SNLI-train & SNLI-contradiction \\ \hline
\multirow{2}{*}{ROCStories} & 1,000 & ROCStories-1st & ROCStories-2nd \\ \cline{2-4}
 & 1,000 & ROCStories-1st & ROCStories-3rd \\ \hline
COPA & 1,000 & COPA-premise & COPA-effect \\ \hline\hline
        Total & 39,093 & - & - \\ \hline
\end{tabular}%
}
\caption{\small JOCI corpus statistics, where \label{tab:joci_stat} each
    subset consists of different sources for context-and-hypothesis pairs,
    each annotated with common-sense ordinal labels.
AGCI-\textsc{wk} represents candidates generated based on world knowledge.
AGCI-\textsc{nn} represents candidates generated via neural methods.}
\end{table}

\subsection{Crowdsourced Ordinal Label Annotation}
\label{sec:crowdsourced_ordinal_annotation}
\begin{figure}[!ht]
    \includegraphics[width=0.45\textwidth]{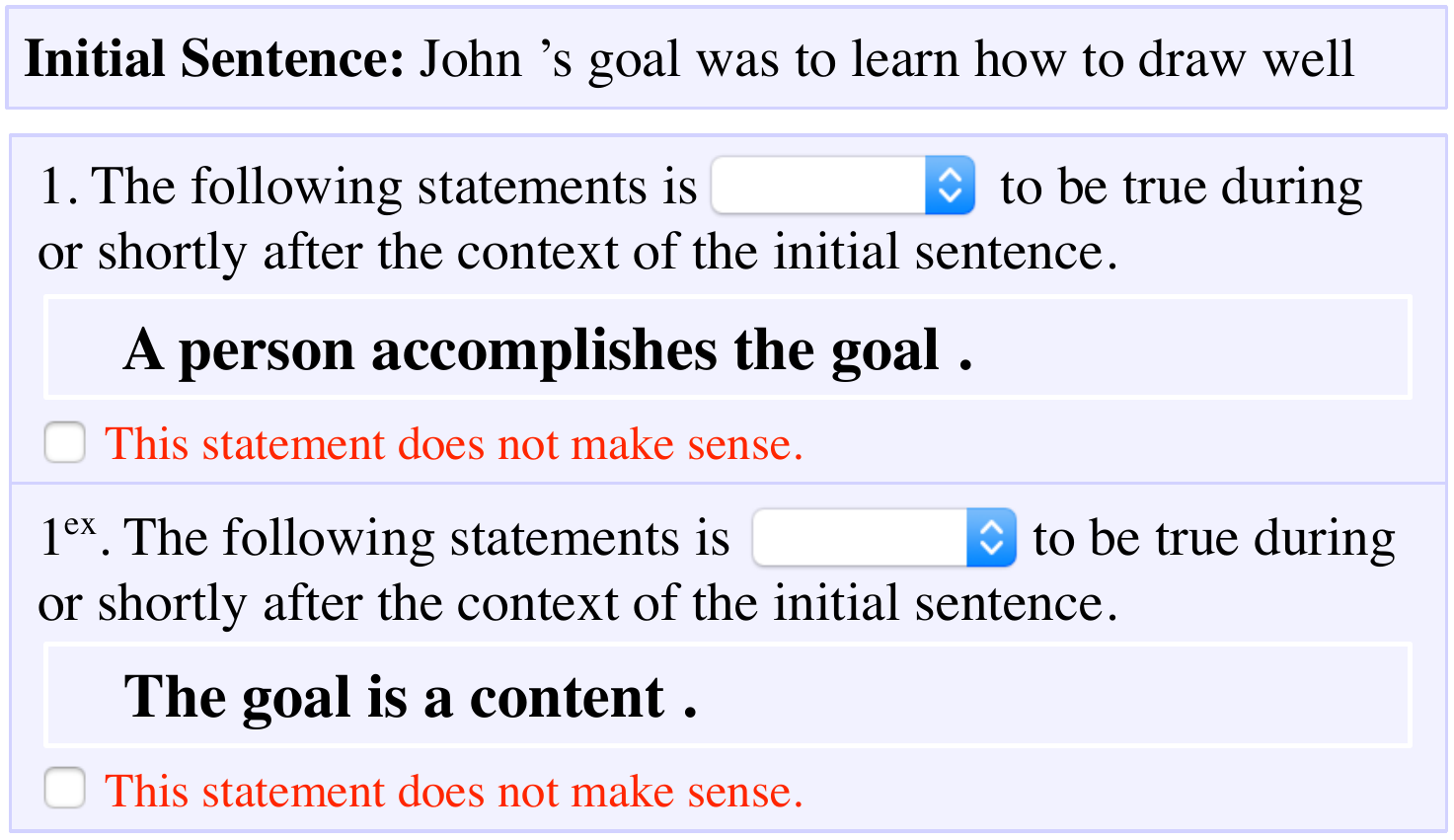}
\centering
\caption{\small The annotation interface, with a drop-down list provides ordinal labels to select.}
\label{fig:hit_example}
\end{figure}

We use Amazon Mechanical Turk to annotate the hypotheses with ordinal labels.
In each HIT (Human Intelligence Task), a worker is presented with one context
and one or two hypotheses, as shown in~\Cref{fig:hit_example}.
First, the annotator sees an ``Initial Sentence" (context),
e.g. ``John's goal was to learn how to draw well.'', and is then asked about
the plausibility of the hypothesis, e.g. ``A person accomplishes the goal''.
In particular, we ask the annotator how plausible the hypothesis is true
during or shortly after, because without this constraint, most sentences are
technically plausible in some imaginary world.

If the hypothesis does not make sense\footnote{``Not making sense" means that
inferences that are incomplete sentences or grammatically wrong.}, the workers
can check the box under the question and skip the ordinal annotation.
In the annotation, about 25\% of hypotheses are marked as not making sense,
and are removed from our data.
\Reviewer{C:  what are the criteria and guidelines for removing the 25\% of
    AGCI that "does not make sense"? Who did this winnowing, and how?}

\begin{table*}[!ht]
\centering
\captionsetup{justification=centering}
\small
\begin{tabular}{|c|c|l|l|c|}
\hline
M. & Labels & \multicolumn{1}{c|}{\bf Context} & \multicolumn{1}{c|}{\bf Hypothesis} \\ \hline
5 & [5,5,5] & \parbox[c][1em][c]{0.44\textwidth}{John was excited to go to the fair} & The fair opens . \\ \hline
4 & [4,4,3] & \parbox[c][1em][c]{0.44\textwidth}{Today my water heater broke} & A person looks for a heater . \\ \hline
3 & [3,3,4] & \parbox[c][1em][c]{0.44\textwidth}{John 's goal was to learn how to draw well} & A person accomplishes the goal . \\ \hline
2 & [2,2,2] & \parbox[c][1em][c]{0.44\textwidth}{Kelly was playing a soccer match for her University} & The University is dismantled . \\ \hline
1 & [1,1,1] & \parbox[c][2.3em][c]{0.44\textwidth}{A brown-haired lady dressed all in blue denim sits in a group of pigeons .} & People are made of the denim . \\ \hline
5 & [5,5,4] &  \parbox[c][2.3em][c]{0.44\textwidth}{Two females are playing rugby on a field, one with a blue uniform and one with a white uniform .} & Two females are play sports outside . \\ \hline
4 & [4,4,3] & \parbox[c][1em][c]{0.44\textwidth}{A group of people have an outside cookout .} & People are having conversations . \\ \hline
3 & [3,3,3] & \parbox[c][1em][c]{0.44\textwidth}{Two dogs fighting, one is black, the other beige .} & The dogs are playing . \\ \hline
2 & [2,2,3] & \parbox[c][3.3em][c]{0.44\textwidth}{A bare headed man wearing a dark blue cassock, sandals, and dark blue socks mounts the stone steps leading into a weathered old building} & A man is in the middle of home building . \\ \hline
1 & [1,1,1] & \parbox[c][2.3em][c]{0.44\textwidth}{A skydiver hangs from the undercarriage of an airplane or some sort of air gliding device} & A camera is using an object . \\ \hline
\end{tabular}%
\caption{\small Examples of context-and-hypothesis pairs with ordinal judgements and Median value. \\
(The upper 5 rows are samples from AGCI-\textsc{wk}. The lower 5 rows are samples from AGCI-\textsc{nn}.)\label{tab:joci_examples}}
\end{table*}

With the sampled contexts and the auto-generated hypotheses, we prepare 
50K common-sense inference examples for crowdsourced annotation in bulk.
In order to guarantee the quality of annotation, we have each example annotated
by three workers. We take the median of the three as the gold label.
\Cref{tab:hit_stat} shows the statistics of the crowdsourced efforts.

\begin{table}[!ht]
\centering
\small
\begin{tabular}{ll}
    \toprule
\# examples & 50,832 \\
\# participated workers & 150 \\
average cost per example & 1.99\cent \\
average work time per example & 20.71s \\
    \bottomrule
\end{tabular}
\caption{\small Statistics of the crowdsourced efforts.\label{tab:hit_stat}}
\end{table}

To make sure non-expert workers have a correct understanding of our task,
before launching the later tasks in bulk, we run two pilots to create a pool
of qualified workers.
In the first pilot, we publish 100 examples.
Each example is annotated by five workers.
From this pilot, we collect a set of ``good" examples which have 100\%
annotation agreement among workers.
The ordinal labels chosen by the workers are regarded as the gold labels.
In the second pilot, we randomly select two ``good" (high-agreement) examples for each ordinal label and publish a HIT
with these examples.
To measure workers' agreement, we calculate the average of
quadratic weighted Cohen's $\kappa$ scores between workers' annotation.
By setting a threshold of the average of $\kappa$ scores to 0.7, we are able
to create a pool that has over 150 qualified workers.
\Reviewer{B: you compute Cohen's kappa to identify reliable
  annotators; however, you use a completely different metric to
  argue that your annotation is reliable (Section 5.3), namely
  standard deviation. I guess this comes down to the question of
  what an appropriate reliability measure for an ordinal scale
  annotation is: please make up your mind, and stick to it. Let
  me say that in my experience, standard deviation is quite
  nonstandard and therefore difficult to compare to other
  annotation tasks. In MT evaluation, for example, Kappa and
  related figures (like Krippendorff's alpha) appear to be the
  standard choices. Using a standard measure here would greatly
  improve the strength of the story.}

\subsection{Corpus Characteristics}

\Reviewer{C:  section 5.3 describes corpus characteristics without saying what sort of
corpus characteristics are desirable, or why. There is no clear claim here.}
We want a corpus with reliable inter-annotator agreement.
Additionally, in order to evaluate or train a common-sense
inference system, we ideally need a corpus that provides for every
ordinal likelihood value as many inference examples as possible.
In this section, we investigate the characteristics of the JOCI corpus.
We also compare JOCI with
related resources under our annotation protocol.


\begin{figure*}[tbp]
    \centering
    \begin{subfigure}[b]{0.32\textwidth}
        \includegraphics[width=\textwidth]{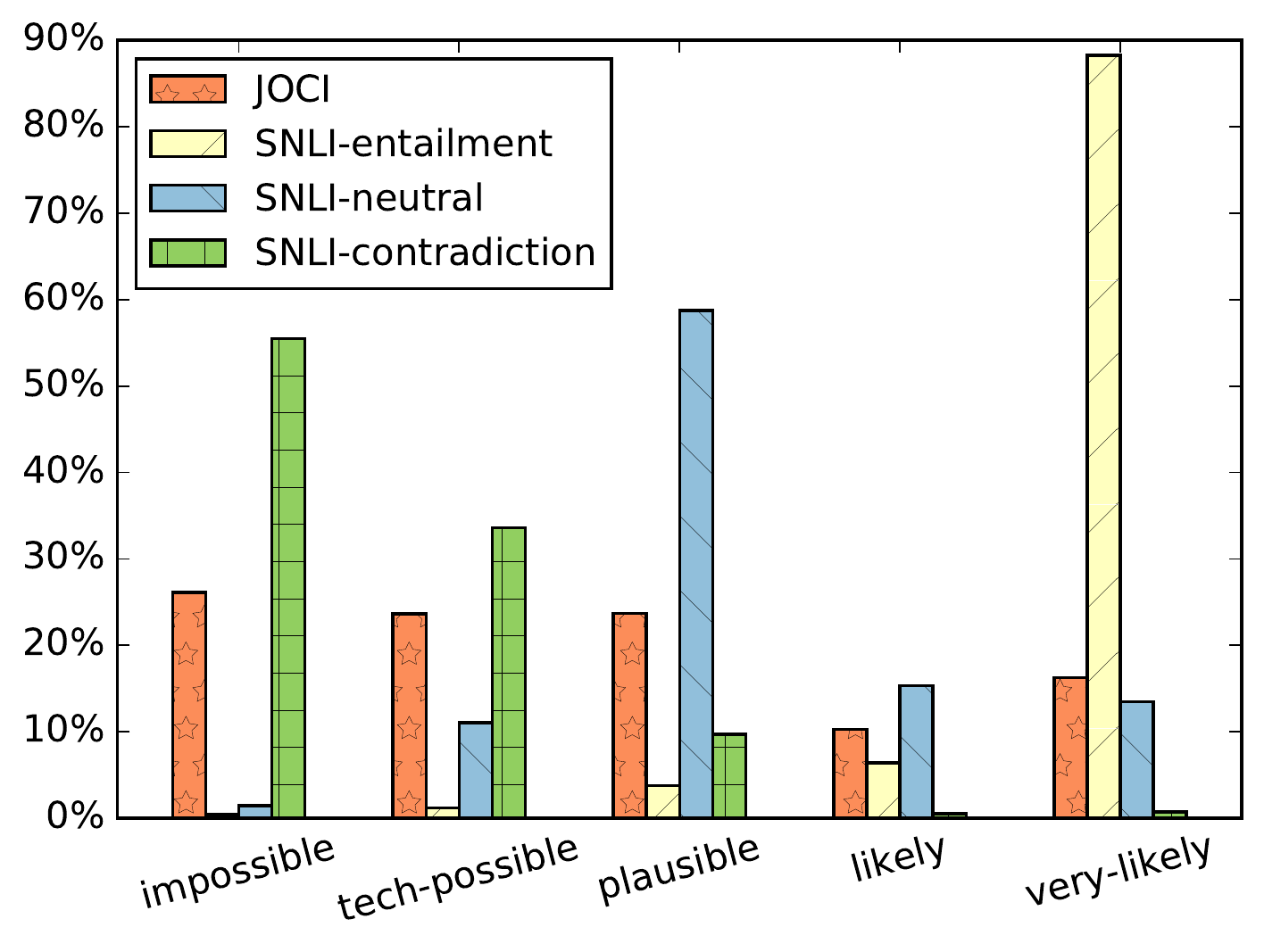}
        \caption{JOCI vs. SNLI}
        \label{fig:snli_dist}
    \end{subfigure}
    \hspace{0em}
    \begin{subfigure}[b]{0.32\textwidth}
        \includegraphics[width=\textwidth]{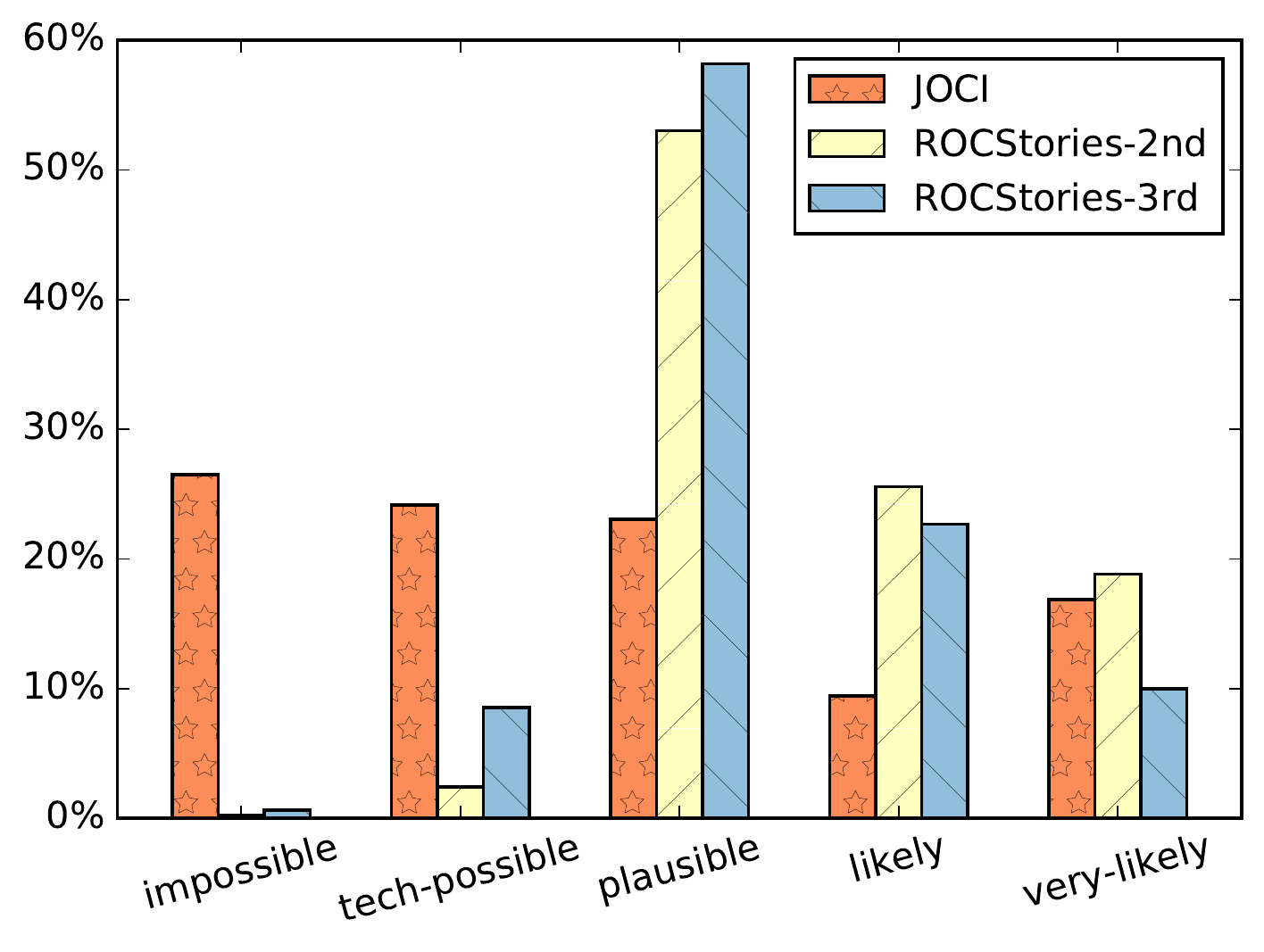}
        \caption{JOCI vs. ROCStories}
        \label{fig:roc_dist}
    \end{subfigure}
    \hspace{0em}
    \begin{subfigure}[b]{0.32\textwidth}
        \includegraphics[width=\textwidth]{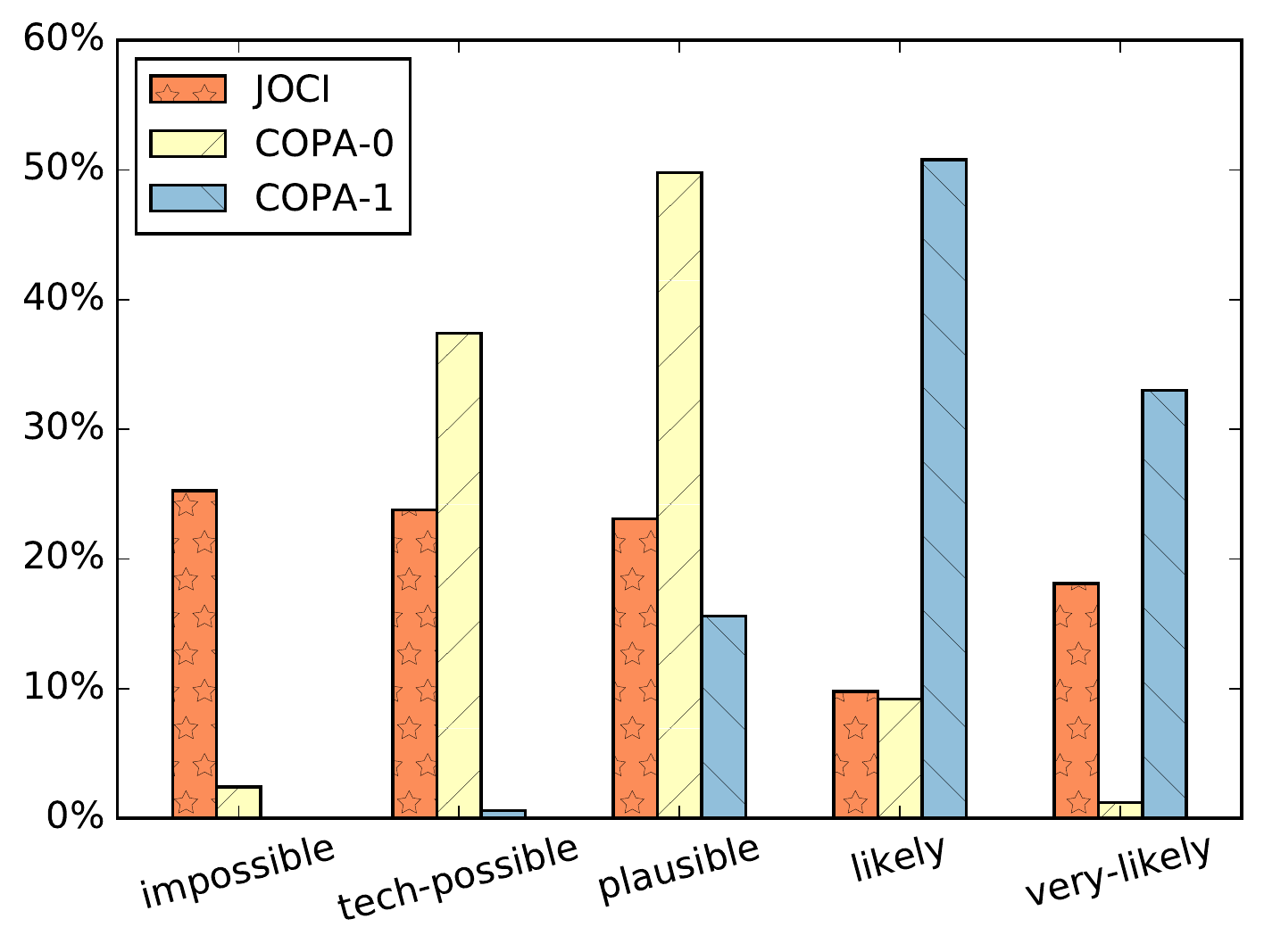}
        \caption{JOCI vs. COPA}
        \label{fig:copa_dist}
    \end{subfigure}
\captionsetup{justification=centering}
\caption{Comparison of normalized distributions between JOCI and other
corpora. \label{fig:all_dist}}
\end{figure*}

\begin{figure}[h!]
    \includegraphics[width=0.43\textwidth]{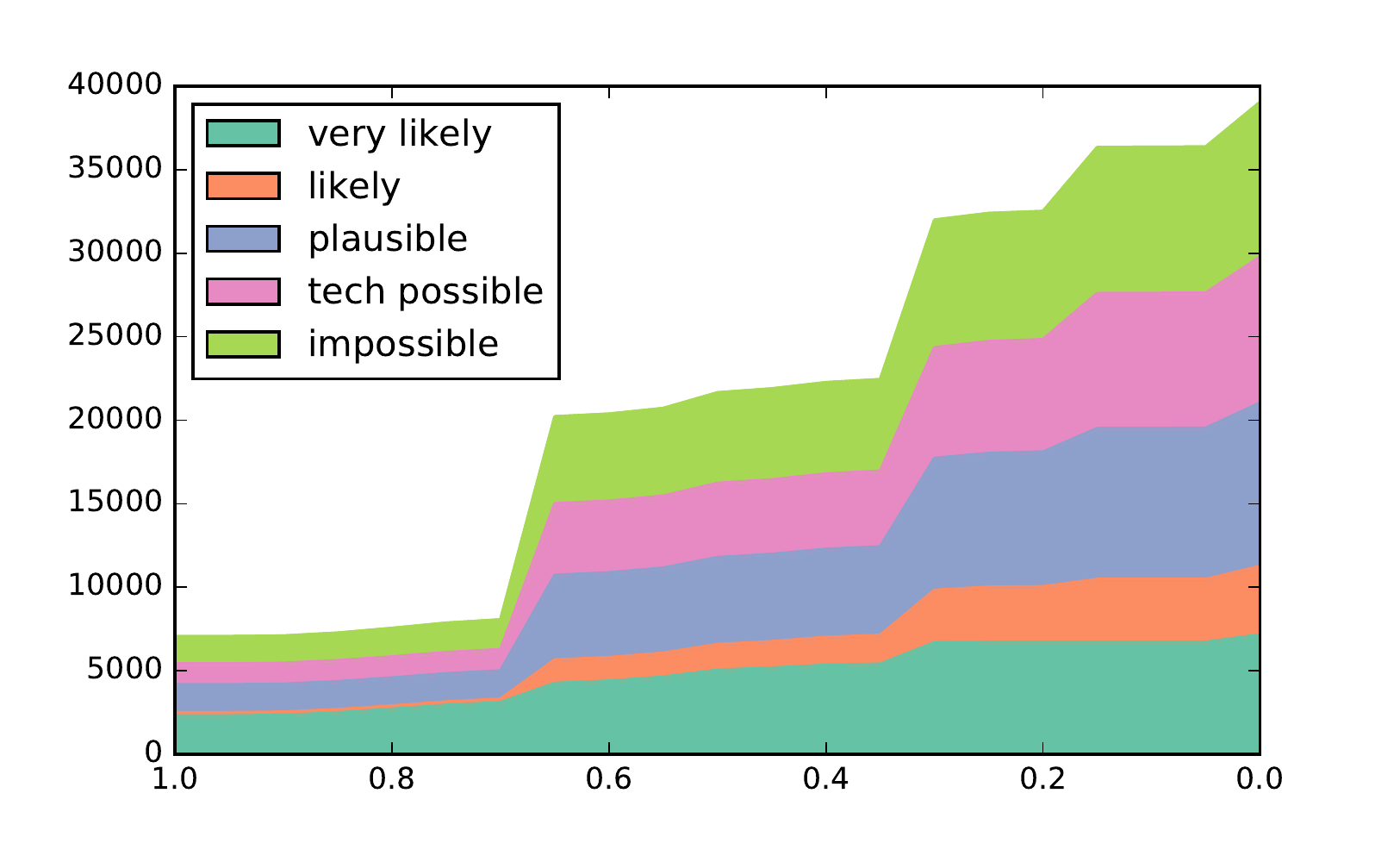}
\centering
\caption{\small Data growth along averaged $\kappa$ scores.
\label{fig:data_growth}}
\end{figure}

\noindent\textbf{Quality:} We measure the quality of each pair by calculating
Cohen's $\kappa$ of workers' annotations.
The average $\kappa$ of the JOCI corpus is 0.54.
\Cref{fig:data_growth} shows the growth of the size of JOCI
as we decrease the threshold of the averaged $\kappa$ to filter pairs.
Even if we place a relatively strict threshold ($>$0.6), we still get
a large subset of JOCI with over 20K pairs.
\Cref{tab:joci_examples} contains pairs randomly sampled from this subset,
qualitatively confirming we can generate and collect annotations of 
pairs at each ordinal category.

\Reviewer{C: The authors do not provide clearly stated method-independent criteria for
types of sentence pair that they would like to see in their corpus, nor do
they provide such criteria for types of inference that they want their
system to be able to handle. This is not necessarily a fault, since it could
be that the corpus which is produced is useful for some downstream task. But
regrettably, the authors do not attempt this, limiting themselves to the
intermediate task of constructing textual inference examples. This IS a
fault, because it means that there is no very clear basis for deciding
whether the newly created textual inference examples are worth having. The
activity reported here may be useful, or not: the work reported here does
not make a strong case either way.}

\noindent\textbf{Label Distribution:} 
We believe datasets with wide support of label distribution are important in
training and evaluating systems to recognize ordinal scale inferences.
\Cref{fig:snli_dist} shows the normalized label distribution of JOCI vs.
SNLI\@. As desired, JOCI covers a wide range of ordinal
likelihoods, with many samples in each ordinal scale. Note also how traditional
RTE labels are related to ordinal labels, although many inferences
in SNLI require no
common-sense knowledge (e.g. paraphrases). As expected, \textit{entailments}
are mostly considered \textit{very likely}; \textit{neutral} inferences 
mostly \textit{plausible}; and \textit{contradictions} likely
to be either \textit{impossible} or \textit{technically
  possible}.




\Cref{fig:roc_dist} shows the normalized distributions of JOCI and
ROCStories.  Compared with ROCStories, JOCI still covers a wider range
of ordinal likelihood.  We observe in ROCStories that, while 2nd
sentences are in general more likely to be true than 3rd, a large
proportion of both 2nd and 3rd sentences are \textit{plausible}, as
compared to \emph{likely} or \emph{very likely}.  This matches
intuition: pragmatics dictates that subsequent sentences in a standard
narrative carry new information.\footnote{I.e., if subsequent
  sentences in a story were always \emph{very likely}, then those
  would be boring tales; the reader could infer the conclusion based
  on the introduction.  While at the same time if most subsequent
  sentences were only technically possible, the reader would give up
  in confusion.}  That our protocol picks this up is an encouraging
sign for our ordinal protocol, as well as suggestive that the makeup
of the elicited ROCStories collection is indeed ``story like.''

\begin{figure}[t!]
    \includegraphics[width=0.43\textwidth]{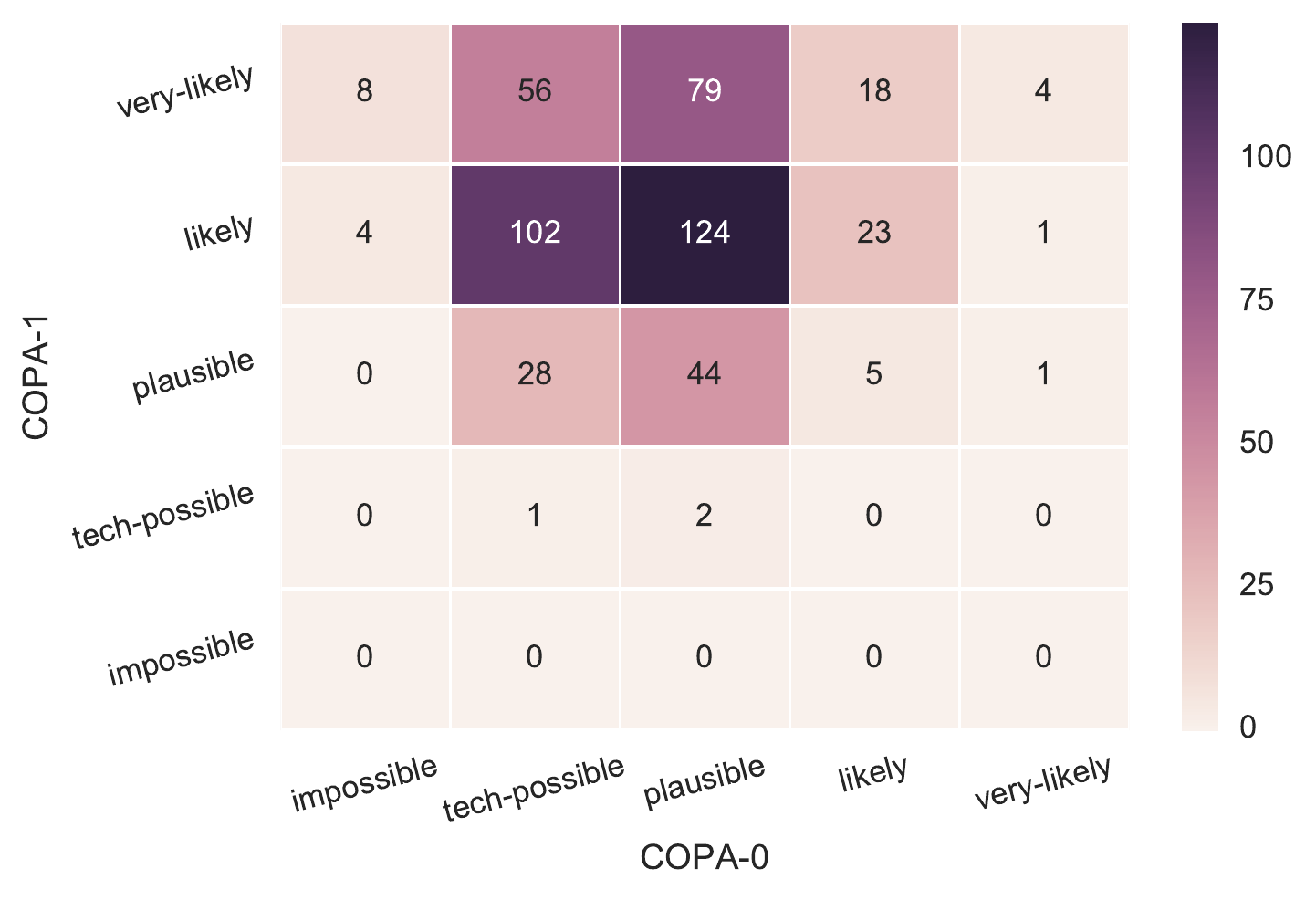}
\centering
\caption{\small COPA heatmap.
\label{fig:copa_heatmap}}
\end{figure}

For the COPA dataset, we make use only of the pairs in which the
alternatives are plausible effects (rather than causes) of the
premise, as our protocol more easily accommodates these
pairs.\footnote{Specifically, we treat premises as contexts and effect
  alternatives as possible hypotheses.} Annotating this section of COPA
with ordinal labels provides an enlightening and validating view of the
dataset.  \Cref{fig:copa_dist} shows the normalized distribution of
COPA next to that of JOCI.  (COPA-1 alternatives are marked as
most plausible; COPA-0 are not.)  True to its name, the majority
of COPA alternatives are labeled as either \textit{plausible} or
\textit{likely}; almost none are \textit{impossible}.  This is
consistent with the idea that the COPA task is to determine which of
two \textit{possible} options is the more plausible.
\Cref{fig:copa_heatmap} shows the joint distribution of ordinal labels on
(COPA-0,COPA-1) pairs.  As expected, the densest areas of the
heatmap lie above the diagonal, indicating that in almost every pair,
COPA-1 received a higher likelihood judgement than COPA-0.

\begin{figure}[t]
\centering
\captionsetup[subfigure]{aboveskip=-3pt,belowskip=-1pt}
\begin{subfigure}{0.4\textwidth}
     \includegraphics[width=\textwidth]{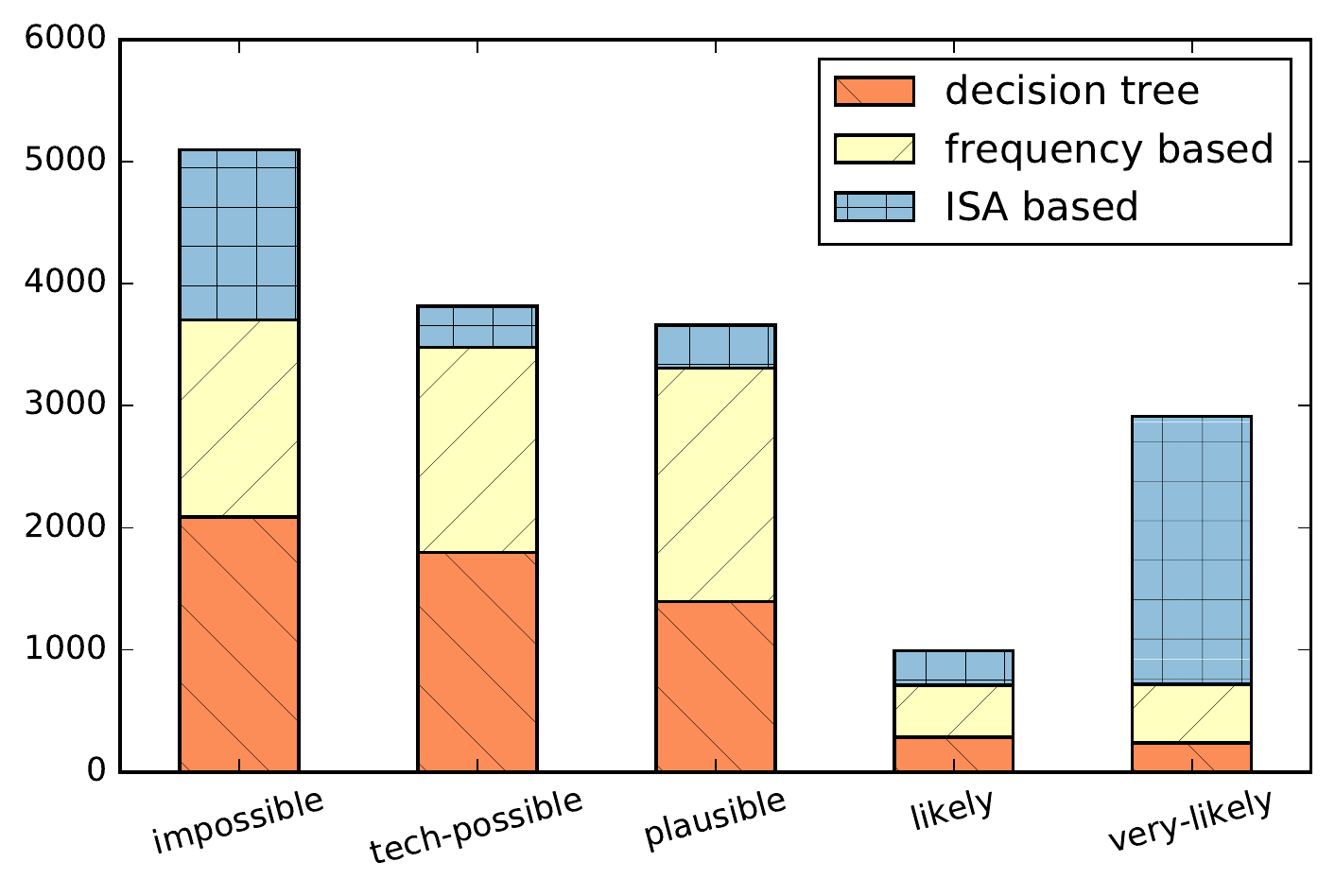}
     \caption{Distribution of AGCI-\textsc{wk}
     \label{fig:agci-wk_dist}}
\end{subfigure}

\begin{subfigure}{0.4\textwidth}
     \includegraphics[width=\textwidth]{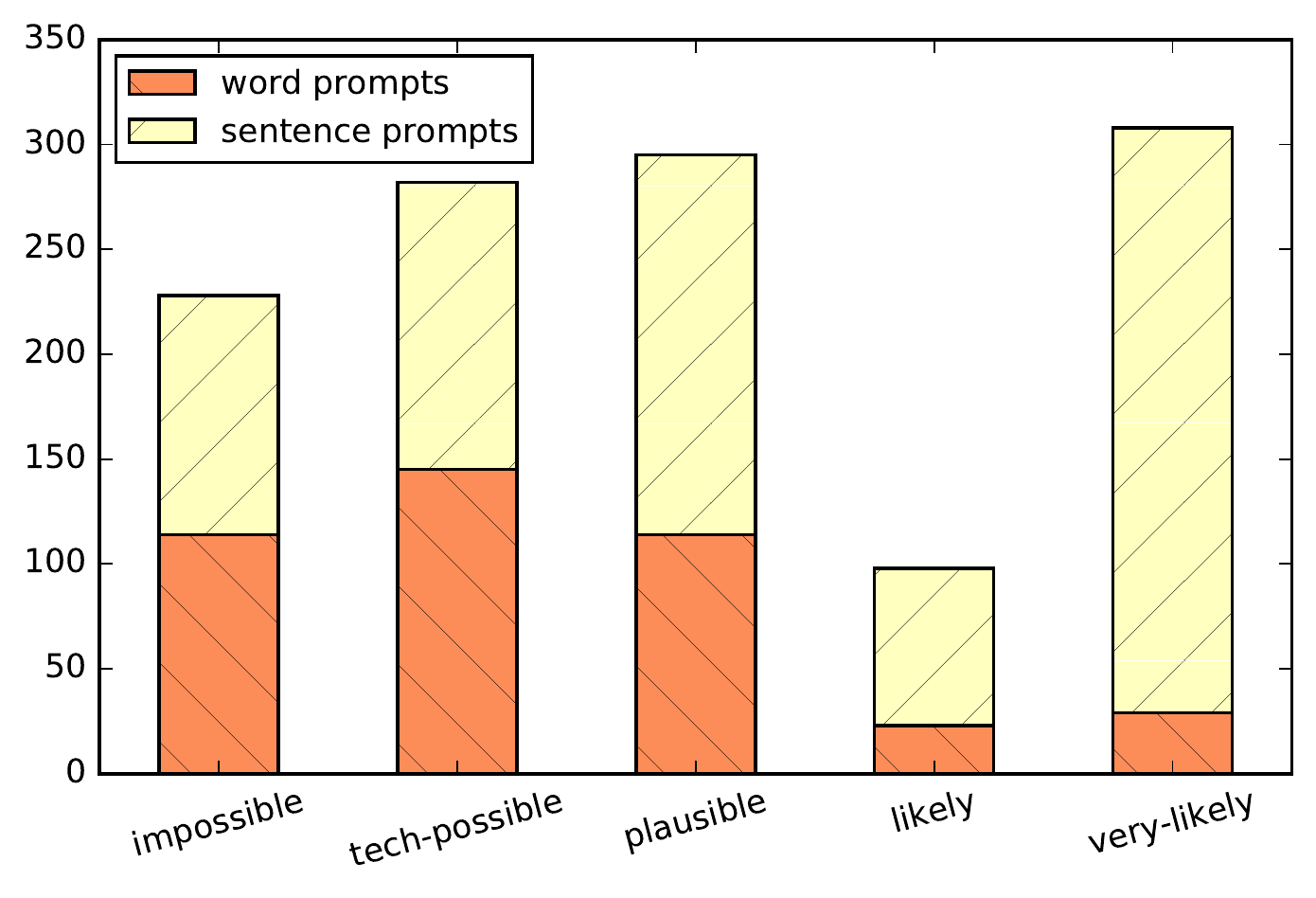}
     \caption{Distribution of AGCI-\textsc{nn}
     \label{fig:agci-nn_dist}}
\end{subfigure}
\caption{Label distributions of AGCI.\label{fig:agci_dist}}
\end{figure}

\noindent\textbf{Automatic Generation Comparisons:}
We compare the label distributions of different methods for automatic generation
of common-sense inference (AGCI) in \Cref{fig:agci_dist}.
Among ACGI-\textsc{wk} (generation based on world knowledge) methods, the ISA
strategy yields a bimodal distribtuion, with the majority of inferences labeled \textit{impossible} or \textit{very likely}.
This is likely because most copular statements generated with the ISA strategy will either be categorically true or false.
In contrast, the decision tree and frequency based strategies generate many
more hypotheses with intermediate ordinal labels.
This suggests the propositional templates (learned from text) capture many
``possibilistic'' hypotheses, which is our aim.

\Reviewer{C: it is no business of the authors to describe the differences between the
AGCI-NN strategies as "interesting". Just delete the word and let the reader
decide.}
The two AGCI-\textsc{nn} (generation via neural methods) strategies show interesting differences in label distribution as well.
Sequence-to-sequence decodings with full-sentence prompts lead to more \textit{very likely} labels than single-word prompts.
The reason may be that the model behaves more similarly to SNLI entailments when it has access to all the information in the context.
When combined, the five AGCI strategies (three AGCI-\textsc{wk} and two AGCI-\textsc{nn}) provide reasonable coverage over all five categories, as can be seen in \Cref{fig:all_dist}.



%


\section{Predicting Ordinal Judgments}
\label{sec:prediction}
\Reviewer{C: section 6 is reasonable if and only if the authors have convinced the
reader that ordinal judgments of the type created are valuable. Otherwise it
leaves the reader wondering what they have learned from the experiments.}

We want to be able to predict ordinal judgments of the kind presented in this corpus.
Our goal in this section is to establish baseline results and explore what kinds of features are useful for predicting ordinal common-sense inference.
To do so, we train and test a logistic ordinal regression model
$g_\theta(\phi({\mathcal C},{\mathcal H}))$, which outputs ordinal labels
using features $\phi$ defined on context-inference pairs.
Here, $g_\theta(\cdot)$ is a regression model with $\theta$ as trained
parameters; we train using the margin-based method of \cite{rennie2005loss},
implemented in \cite{pedregosa2015feature},\footnote{LogisticSE:
\url{http://github.com/fabianp/mord}} with the following features:



\noindent\textbf{Bag of words features (\textsc{bow})}: We compute (1)
``BOW overlap'' (size of word overlap in $\mathcal{C}$ and $\mathcal{H}$), and
(2) BOW overlap divided by the length of $\mathcal{H}$.

\noindent\textbf{Similarity features (\textsc{sim})}: Using Google's word2vec
vectors trained on 100 billion tokens of GoogleNews,\footnote{The GoogleNews
embeddings are available at: \url{https://code.google.com/archive/p/word2vec/}}
we (1) sum the vectors
in both the context and hypothesis and compute the cosine-similarity of the
resulting two vectors (``similarity of average''), and (2) compute the
cosine-similarity of all word pairs across the context and inference, then
average those similarities (``average of similarity'').




\noindent\textbf{Seq2seq score features (\textsc{s2s})}: We compute the log
probability $\log P(\mathcal{H} | \mathcal{C})$ under the
sequence-to-sequence model described in \cref{subsec:neural}.
There are five variants: (1) Seq2seq trained on SNLI ``entailment'' pairs
only, (2) ``neutral'' pairs only, (3) ``contradiction'' pairs only,
(4) ``neutral'' and ``contradiction'' pairs, and (5) SNLI pairs (any label)
with the context (premise) replaced by an empty string.


\noindent\textbf{Seq2seq binary features (\textsc{s2s-bin})}: Binary
indicator features for each of the five seq2seq model variants, indicating
that model achieved the lowest score on the context-hypothesis pair.


\noindent\textbf{Length features (\textsc{len})}: This set comprises three
features: the length of the context (in tokens), the difference in length
between the context and hypothesis, and a binary feature indicating if the
hypothesis is longer than the context.

\subsection{Analysis}


\Reviewer{D: the automatic generation of examples via neural methods is not as
well motivated as the first approach, and more analysis is needed to compare
and contrast the nature of examples generated using each of these
approaches. For instance, it would be interesting to complement Table 9 with
further results using examples generated using Gigaword sentences (AGCI-WK)
vs. neural network generated sentences (AGCI-NN). Do artificial sentences
yield easier examples?  Figure 8 suggests that those generated from single
word prompts are mostly impossible or technically possible, and I wonder to
what extent they are simply non-sensical sentences that might be easier to
detect.}
We train and test our regression model on two subsets of the JOCI corpus,
which, for brevity, we call ``A'' and ``B.''
``A'' consists of 2,976 sentence pairs (i.e., context-hypothesis pairs) from
SNLI-train annotated with ordinal labels. This corresponds to the three rows
labeled SNLI in Table \ref{tab:joci_stat} ($993+988+995=2,976$ pairs), and
can be viewed as a textual entailment dataset re-labeled with ordinal judgments.
``B'' consists of 6,375 context-inference pairs, in which the contexts are
the same 2,976 SNLI-train premises as ``A'', and the hypotheses are generated
based on world knowledge (\S\ref{sec:gwke}); these pairs are also annotated
with ordinal labels. This corresponds to a subset of the row labeled AGCI in
Table \ref{tab:joci_stat}. A key difference between ``A'' and ``B'' is that
the hypotheses in ``A'' are human-elicited, while those in ``B'' are
auto-generated; we are interested in seeing whether this affects the task's
difficulty.\footnote{Details of the data split is reported in the dataset release.}




\begin{table}[h!]
\begin{centering}
\small
\begin{tabular}{l@{\hspace{3.5mm}}c@{\hspace{2mm}}@{\hspace{2mm}}c@{\hspace{2mm}}@{\hspace{2mm}}c@{\hspace{2mm}}@{\hspace{2mm}}c@{}}
    \toprule
Model          & A-train & A-test & B-train & B-test \\
\hline
Regression: $g_\theta(\cdot)$ &\textbf{2.05}&\textbf{1.96}&\textbf{2.48}&\textbf{2.74} \\
Most Frequent             & 5.70    & 5.56   & 6.55    & 7.00 \\
Freq. Sampling            & 4.62    & 4.29   & 5.61    & 5.54 \\
Rounded Average            & 2.46    & 2.39   & 2.79    & 2.89 \\
One-vs-All            & 3.74    & 3.80   & 5.14    & 5.71 \\
    \bottomrule
\end{tabular}
\caption{\small Mean squared error.
\label{fig:mse}}
\end{centering}
\end{table}

\begin{table}[h!]
\begin{centering}
\small
\begin{tabular}{l@{\hspace{3.5mm}}c@{\hspace{2mm}}@{\hspace{2mm}}c@{\hspace{2mm}}@{\hspace{2mm}}c@{\hspace{2mm}}@{\hspace{2mm}}c@{}}
    \toprule
Model          & A-train & A-test & B-train & B-test \\
\hline
Regression: $g_\theta(\cdot)$ &\textbf{.39*}&\textbf{.40*}&\textbf{.32*}&\textbf{.27*} \\
Most Frequent             & .00*    & .00*   & .00*     & .00* \\
Freq. Sampling            & .03\ph  & .10\ph & .01\ph   & .01\ph \\
Rounded Average           & .00*    & .00*   & .00*     & .00* \\
One-vs-All                & .31*    & .30*   & .28*     & .24* \\
    \bottomrule
\end{tabular}
\caption{\small Spearman's $\rho$. (*p-value$<$.01)
\label{fig:spearman}}
\end{centering}
\end{table}

\Reviewer{B: Similarly, the evaluation metric that you report on your
  baseline models (Figure 9) is highly nonstandard.
  You cite Textual Similarity in Section 2, and I believe that
  as an established prediction task on an ordinal scale the
  evaluation metrics established there (acc, p, r, f1) might
  make a good model for your task. At least if you do not want
  to adopt them, you should have a very good reason for doing so.}

Tables \ref{fig:mse} and \ref{fig:spearman} show each model's performance (mean squared error and Spearman's $\rho$, respectively) in predicting ordinal labels.\footnote{MSE
    and Spearman's $\rho$ are both commonly used evaluations in ordinal prediction tasks
\shortcite{baccianella2009evaluation,bennett2007netflix,gaudette2009evaluation,agresti2003categorical,popescu2009comparing,liu2015learning,gella2013unsupervised}.}
We compare our ordinal regression model $g_\theta(\cdot)$ with these baselines:

\textbf{Most Frequent}: Select the ordinal class appearing most often in train.

\textbf{Frequency Sampling}: Select an ordinal label according to their distribution in train.

\textbf{Rounded Average}: Average over all labels from train rounded to nearest ordinal.

\textbf{One-vs-All}: Train one SVM classifier per ordinal class and select the class label with the largest corresponding margin. We train this model with the same set of features as the ordinal regression model.

Overall, the regression model achieves the lowest MSE and highest $\rho$,
implying that this dataset is learnable and tractable. Naturally, we would
desire a model that achieves MSE under 1.0, and we hope that the release of
our dataset will encourage more concerted effort in this common-sense inference task.
Importantly, note that performance on A-test is better than on B-test. We
believe ``B'' is a more challenging dataset because auto-generation of
\text{hypothesis} leads to wider variety than elicitation.





\begin{table}[h!]
\begin{centering}
\small
\begin{tabular}{lcccc}
\toprule
 & \multicolumn{2}{c}{MSE} & \multicolumn{2}{c}{Spear. $\rho$}\\
Feature Set                      & A & B &A &B  \\
\hline
\textsc{all}                 & \textbf{1.96}& \textbf{2.74} &\textbf{.40}*&\textbf{.27}*\\
\textsc{all -- \{sim\}}              & 2.10    & 2.75  & .34*    & .25*\\
\textsc{all -- \{bow\}}              & 2.02    & 2.77  & .37*    & .25*\\
\textsc{all -- \{sim,bow\}}          & 2.31    & 2.79  & .16*    & .20*\\
\textsc{all -- \{s2s\}}              & 2.00    & 2.85  & .38*    & .22*\\
\textsc{all -- \{s2s-bin\}}          & 1.97    & 2.76  & .40*    & .26*\\
\textsc{all -- \{s2s,s2s-bin\}}      & 2.06    & 2.87  & .35*    & .21*\\
\textsc{all -- \{len\}}              & 2.01    & 2.77  & .39*    & .25*\\
\textsc{$\varnothing$ + \{sim\}}     & 2.06    & 3.04  & .35*    & .10\ph\\
\textsc{$\varnothing$ + \{bow\}}     & 2.10    & 2.89  & .34*    & .12*\\
\textsc{$\varnothing$ + \{s2s\}}     & 2.33    & 2.80  & .14\ph  & .20*\\
\textsc{$\varnothing$ + \{s2s-bin\}} & 2.39    & 2.89  & .00\ph  & .00\ph\\
\textsc{$\varnothing$ + \{len\}}     & 2.39    & 2.89  & .00\ph  & .05\ph\\
\bottomrule
\end{tabular}
\caption{\small Ablation results for ordinal regression model on A-test and B-test. (*p-value$<$.01 for $\rho$)
\label{fig:ablation}}
\end{centering}
\end{table}

We also run a feature ablation test.
\Cref{fig:ablation} shows that the most useful features differ for A-test and B-test.
On A-test, where the inferences are elicited from humans, removal of similarity- and bow-based features together results in the largest performance drop.
On B-test, by contrast, removing similarity and bow features results in a comparable performance drop to removing seq2seq features.
These observations point to statistical differences between human-elicited
and auto-generated hypotheses, a motivating point of the JOCI corpus.






\section{Conclusions and Future Work}

In motivating the need for automatically building collections of
common-sense knowledge, \newcite{clark2003knowledge} wrote:

\begin{small}
\begin{quote}
  \emph{``China launched a meteorological satellite into orbit
    Wednesday.'' suggests to a human reader that (among other things)
    there was a rocket launch; China probably owns the satellite;
    the satellite is for monitoring weather; the orbit is around
    Earth; etc}
\end{quote}
\end{small}

The use of \textit{``etc''} summarizes an infinite number of other statements
that a human reader would find to be very likely, likely, technically
plausible, or impossible, given the provided context.

Preferably we could build systems that would automatically learn
common-sense exclusively from available corpora; extracting not just
statements about what is possible, but also the associated
probabilities of how likely certain things are to obtain in any given
context.  We are unaware of existing work that has demonstrated this
to be feasible.


We have thus described a multi-stage approach to common-sense
textual inference: we first extract large numbers of possible
statements from a corpus, and use those statements to generate
contextually grounded context-hypothesis pairs.  These are presented to
humans for direct assessment of subjective likelihood, rather than
relying on corpus data alone.  As the data is automatically generated,
we seek to bypass issues in human elicitation bias.  Further, since
subjective likelihood judgments are not difficult for humans, our
crowdsourcing technique is both inexpensive and scalable.

\Reviewer{us:\emph{The resulting Johns Hopkins Ordinal
Common-sense Inference (JOCI) dataset, and associated novel task
description, is a refinement over prior versions of textual inference,
and captures various levels of speculative common-sense inferences.}
C: There is no convincing data or argument to support the
  claim that this is a refinement over previous versions of textual
  inference. I would be comfortable with the claim that it is a
  possibly interesting ALTERNATIVE to the prior versions, but not that
  it is a refinement.}

Future work will extend our techniques for forward inference
generation, further scale up the annotation of additional examples,
and explore the use of larger, more complex contexts. The resulting
JOCI corpus will be used to improve algorithms for natural language 
inference tasks such as storytelling and story understanding.

\section*{Acknowledgments}
Thank you to action editor Mark Steedman and the anonymous reviewers
for their feedback, as well as colleagues including Lenhart Schubert,
Kyle Rawlins, Aaron White, and Keisuke Sakaguchi.  This
work was supported in part by DARPA LORELEI, the National Science Foundation
Graduate Research Fellowship , and the JHU Human Language Technology Center
of Excellence (HLTCOE).


\bibliography{tacl}
\bibliographystyle{tacl}

\end{document}